\documentclass[10pt,twocolumn,letterpaper]{article}

\usepackage{cvpr}              

\usepackage{algorithm}
\usepackage{algorithmic}
\usepackage{multirow}

\definecolor{cvprblue}{rgb}{0.21,0.49,0.74}
\usepackage[pagebackref,breaklinks,colorlinks,allcolors=cvprblue]{hyperref}

\def\paperID{42583} 
\def\confName{CVPR}
\def\confYear{2026}

\title{Look Before You Fuse: 2D-Guided Cross-Modal Alignment for Robust 3D Detection}

\author{
Xiang Li\textsuperscript{1,2} \quad
Zhangchi Hu\textsuperscript{2} \quad
Xu Xiao\textsuperscript{1,2} \quad
Bin Kong\textsuperscript{1$\dagger$} \\
\textsuperscript{1}Institute of Intelligent Machines, Hefei Institutes of Physical Science, Chinese Academy of Sciences \\
\textsuperscript{2}University of Science and Technology of China \\
{\tt\small \{xiangli, HuZhangchi, xiao\_xu\}@mail.ustc.edu.cn, bkong@iim.ac.cn}
}

\begin{document}

\twocolumn[{
\renewcommand\twocolumn[1][]{#1}
\maketitle
\begin{center}
    \captionsetup{type=figure} 
    \includegraphics[width=1\textwidth]{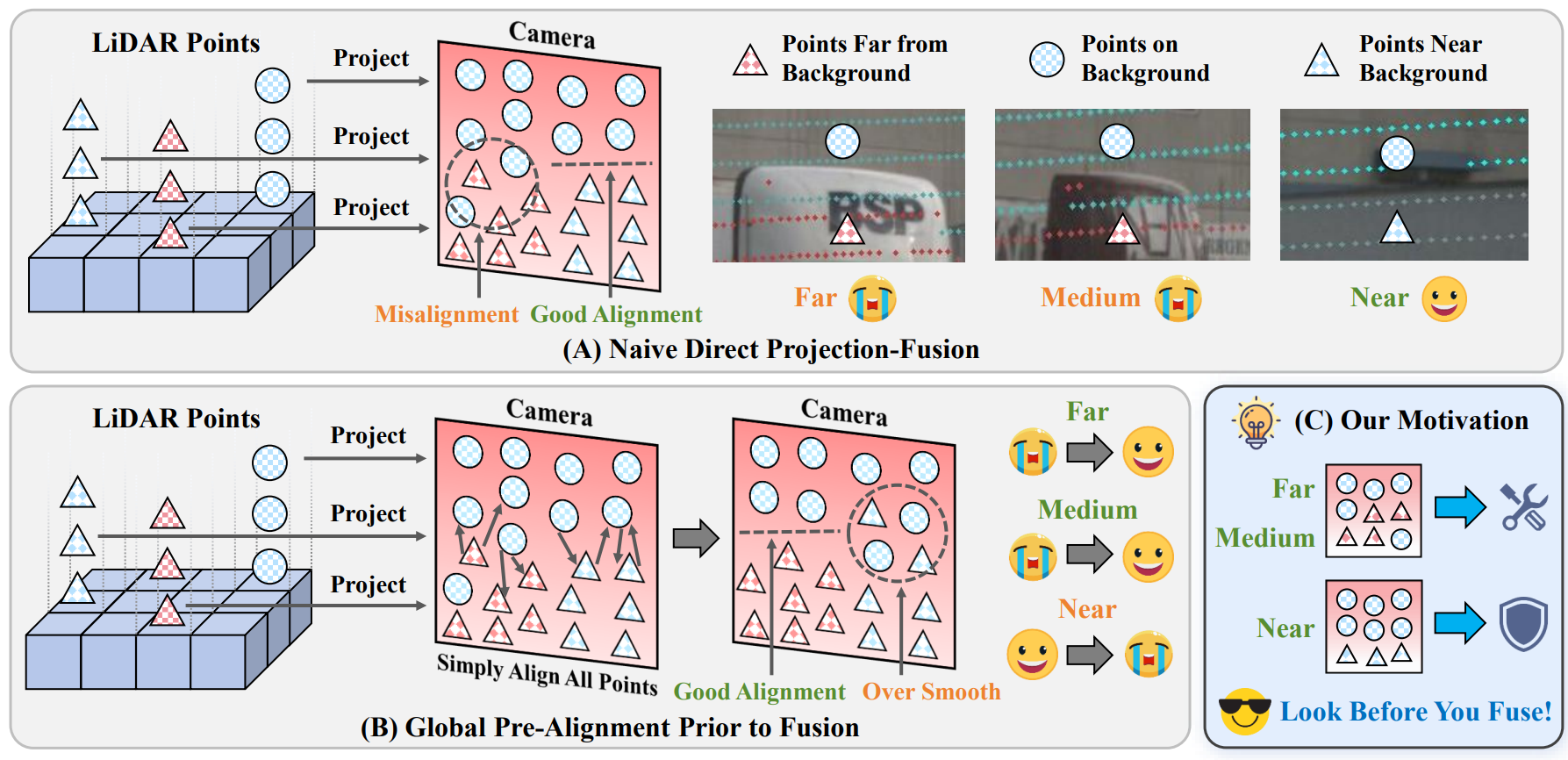} 
    \captionof{figure}{LiDAR points from the distant wall (cyan) are incorrectly projected onto the foreground vehicle (which should appear red) due to a sharp depth change. In contrast, the boundary between background objects like the wall (light cyan) and garage (dark cyan) is correctly projected thanks to a gradual depth transition. This shows that misalignment is most severe at abrupt foreground-background boundaries. And our motivation is to rectify misaligned points while maintaining already aligned points.}
    \label{fig1}
\end{center}
}]

{\let\thefootnote\relax\footnotetext{$^{\dagger}$ Corresponding author.}}
\begin{abstract}
Integrating LiDAR and camera inputs into a unified Bird’s-Eye-View (BEV) representation is crucial for enhancing 3D perception capabilities of autonomous vehicles. However, existing methods suffer from spatial misalignment between LiDAR and camera features, which causes inaccurate depth supervision in camera branch and erroneous fusion during cross-modal feature aggregation. The root cause of this misalignment lies in projection errors, stemming from calibration inaccuracies and rolling shutter effect.
The key insight of this work is that locations of these projection errors are not random but highly predictable, as they are concentrated at object-background boundaries which 2D detectors can reliably identify. Based on this, our main motivation is to utilize 2D object priors to pre-align cross-modal features before fusion. To address local misalignment, we propose \textbf{Prior Guided Depth Calibration (PGDC)}, which leverages 2D priors to alleviate misalignment and preserve correct cross-modal feature pairs. To resolve global misalignment, we introduce \textbf{Discontinuity Aware Geometric Fusion (DAGF)} to suppress residual noise from PGDC and explicitly enhance sharp depth transitions at object-background boundaries, yielding a structurally aware representation. To effectively utilize these aligned representations, we incorporate \textbf{Structural Guidance Depth Modulator (SGDM)}, using a gated attention mechanism to efficiently fuse aligned depth and image features. Our method achieves SOTA performance on nuScenes validation dataset, with its mAP and NDS reaching 71.5\% and 73.6\% respectively.
Additionally, on the Argoverse 2 validation set, we achieve a competitive mAP of 41.7\%.
\end{abstract}

\section{Introduction}

Robust 3D perception is fundamental to autonomous driving, where effective sensor fusion is essential as different modalities offer complementary strengths. Images provide rich semantic information but lack accurate depth, while point clouds offer precise geometry and depth but are sparse and lack semantic context. Harnessing these complementary strengths while mitigating their limitations is key to building reliable perception systems.

To fully leverage complementary information from camera and LiDAR, current architectures either use LiDAR for explicit supervision of the 2D-to-3D transformation process \cite{li2023bevdepth}, or integrate two complementary information sources by fusing LiDAR and camera BEV representations \cite{liu2023bevfusion,liang2022bevfusion}. While effective, these methods are fundamentally challenged by the inherent spatial misalignment between sensors \cite{yu2023benchmarking}. Although the performance of these methods has a relatively high theoretical upper bound, reaching this bound requires perfect sensor alignment, which is frequently violated by projection errors. This discrepancy leads to two critical issues. First, it corrupts depth supervision signal, providing noisy or incorrect depth labels to image branch. Second, during cross-modal aggregation, spatial misalignment causes the fusion module to associate semantically mismatched image and geometric features, thereby degrading quality and reliability of the final fused representation.

Existing approaches have attempted to mitigate cross-modal misalignment, yet each suffers from significant drawbacks.
To ensure feature consistency, methods like transfusion \cite{bai2022transfusion} incorporate attention mechanism to query features of a specific modality, avoiding direct projection errors. While this effectively sidesteps the main cause of misalignment, it comes at the cost of sacrificing crucial contextual information.
Other approaches like MetaBEV \cite{ge2023metabev} and RobBEV \cite{wang2024towards} attempt to mitigate misalignment effects by designing more robust and adaptive fusion modules. Although these modules are more resilient to inconsistent inputs, they cannot correct the initial geometric errors from the 2D-3D view transformation. In essence, they are skillfully fusing features that have already been misplaced. Finally, global alignment techniques like GraphBEV \cite{song2024graphbev} directly address the geometric problem, effectively eliminating misalignment in areas with steep depth gradients. However, they tend to unnecessarily smooth geometrically stable regions where misalignment is negligible or even entirely absent, thereby incorrectly modifying already-correct depth values, as illustrated in \cref{fig1}(B).

Prevailing methods often overlook that misalignment is not randomly distributed, but originates from two primary sources: extrinsic calibration errors and motion-induced distortions. The resulting geometric projection errors are depth-dependent, being negligible for nearby objects but significantly exacerbated at greater depths. This depth-dependent distortion creates the most severe feature misalignment at boundaries between foreground objects and their backgrounds. Our approach is built upon the core insight that critical misalignments are concentrated at foreground-background boundaries where sharp discontinuities in depth occur, as illustrated in \cref{fig1}(A), and we can leverage robust 2D object priors to identify these specific regions, locating real misalignment and accurately solving it while maintaining already aligned regions, as illustrated in \cref{fig1}(C).

In this paper, we introduce three synergistic modules. First, to address the misalignment problem at its root, we propose \textbf{(a) Prior Guided Depth Calibration (PGDC)}. Guided by the principle of ``Look Before You Fuse'', our approach posits that rather than mitigating the effects of fused, misaligned data, it is more effective to proactively correct geometric inconsistencies using high-level semantic guidance before the fusion stage. PGDC actively uses 2D detection proposals as explicit geometric priors to locate and correct misaligned point cloud data, providing a significantly more accurate depth map. However, a corrected sparse map alone is not enough. To enable the network to fully understand the scene's geometric structure, we introduce \textbf{(b) Discontinuity Aware Geometric Fusion (DAGF)}. 
DAGF refines the PGDC-calibrated depth by masking points that exhibit large deviations from the raw depth map, subsequently replacing them with more reliable estimates, fitting the regions of sharp depth transition to the object-background boundaries. This self-correction mechanism corrects remaining misalignments when 2D priors are accurate, and reverts PGDC-induced over-smoothing when 2D priors are flawed. Eventually, DAGF produces a representation of the global depth structure. To effectively utilize these calibrated signals for view transformation, we introduce the \textbf{(c) Structural Guidance Depth Modulator (SGDM)}. This module intelligently fuses image features and dense geometric representation from DAGF using a gated attention mechanism. Its purpose is to predict a highly accurate depth distribution for each pixel, enabling a more precise projection of features into the final BEV space.
\begin{figure*}[t]
\centering
\includegraphics[width=1\textwidth]{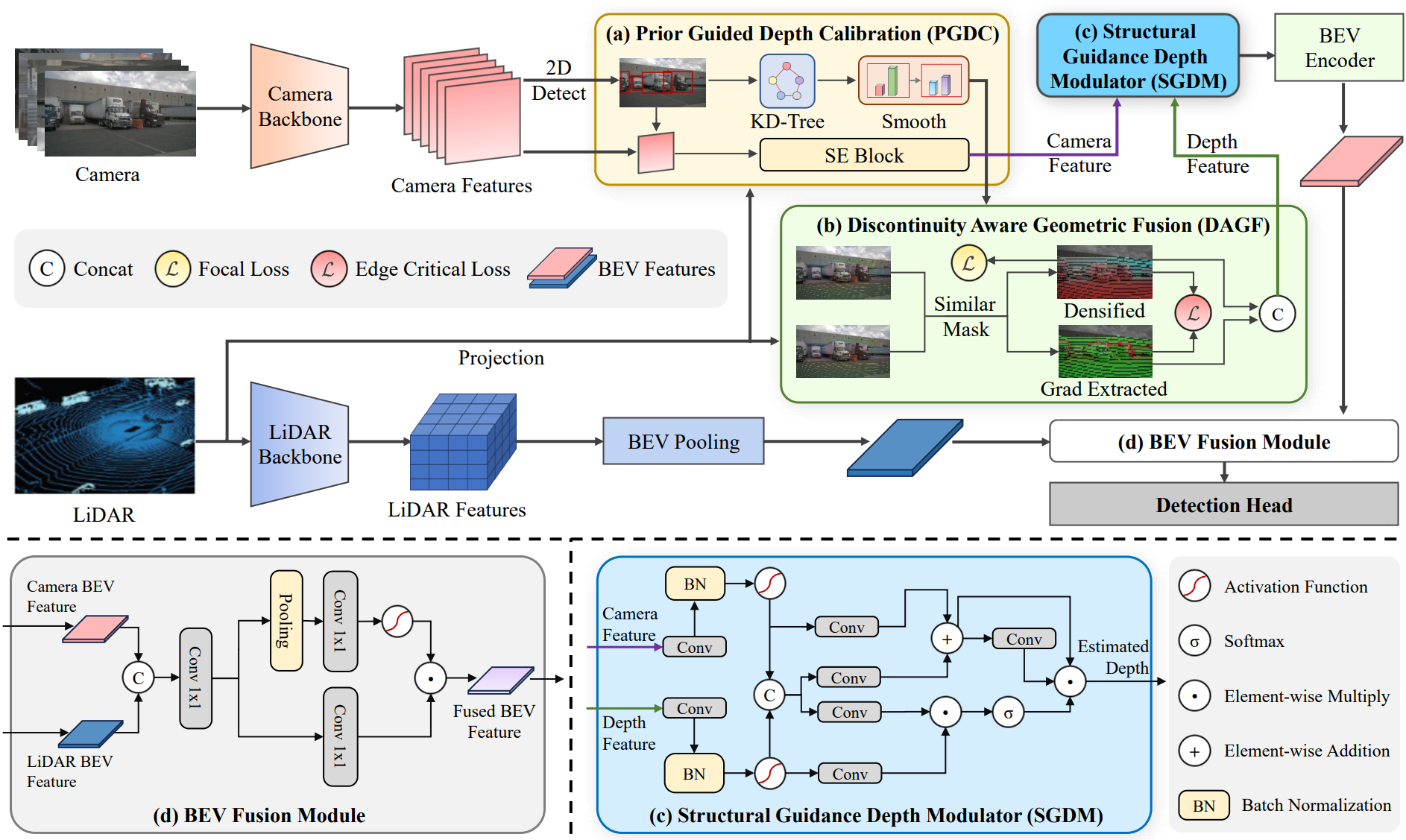} 
\caption{Overview of our proposed framework. (a) Prior Guided Depth Calibration (PGDC) and (b) Discontinuity Aware Geometric Fusion (DAGF), proactively mitigate multi-sensor feature misalignment before view transformation. And (c) Structural Guidance Depth Modulator (SGDM) intelligently fuses image features and dense geometric representation, predicting an accurate depth distribution. Finally, fusing rectified camera BEV features with LiDAR BEV features leads to robust 3D detection.}
\label{fig2}
\end{figure*}

The most significant contributions stemming from our work are summarized below:
\begin{itemize}
    \item We reveal that feature misalignment predominantly occurs at object-background boundaries and propose \textbf{Prior Guided Depth Calibration (PGDC)}, a novel module that actively uses 2D priors to locate and resolve such misalignment, thereby providing more accurate depth information. To the best of our knowledge, we are among the first to explicitly use 2D object detection priors to locally calibrate LiDAR-camera alignment errors before fusion, with a focus on preserving depth discontinuities.
    \item Capitalizing on the intrinsic correlation between depth discontinuities and object-background boundaries, we introduce the \textbf{Discontinuity Aware Geometric Fusion (DAGF)} module that explicitly optimizes cross-modal feature alignment through discontinuity-aware mechanisms, significantly improving geometric consistency in multi-sensor perception.
    \item We introduce the \textbf{Structural Guidance Depth Modulator (SGDM)}, an efficient attention-based module that fuses calibrated visual and geometric cues to generate an accurate depth distribution for view transformation.
    \item Extensive experiments are conducted on nuScenes Dataset, and our method achieves state-of-the-art performance on nuScenes validation dataset with mAP and NDS of 71.5\% and 73.6\%. Additionally, we validate the robustness and generalization of our framework on the Argoverse 2 validation set, achieving a competitive mAP of 41.7\%.
\end{itemize}

\begin{figure*}[t]
\centering
\includegraphics[width=1\textwidth]{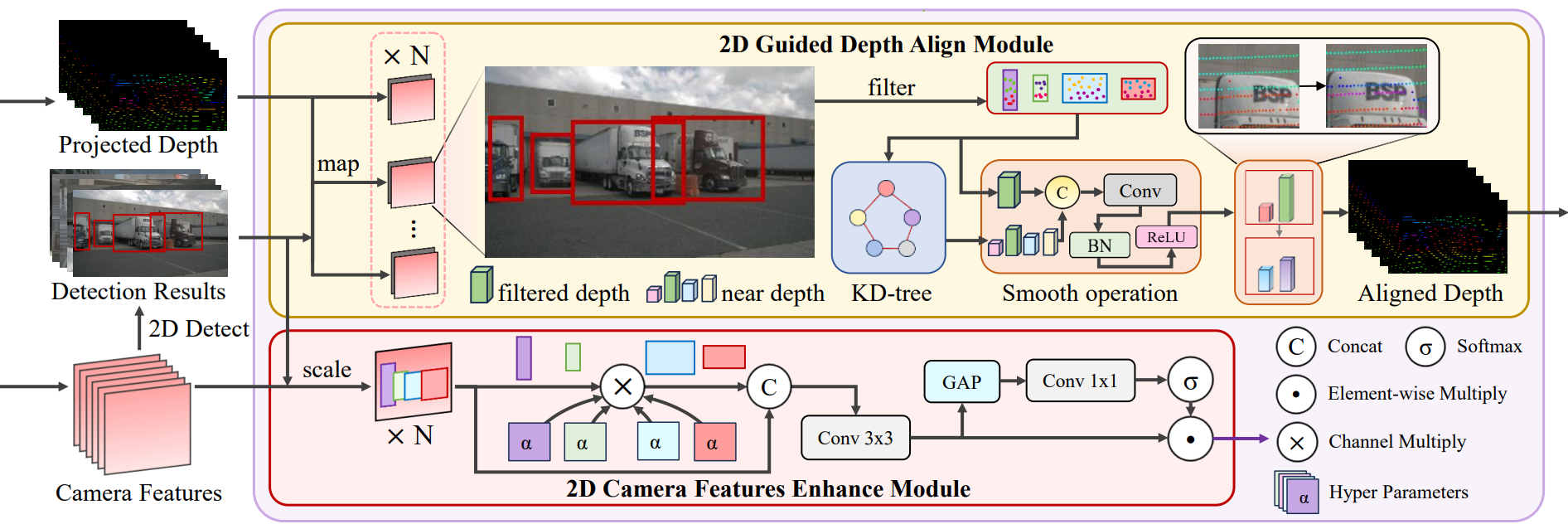} 
\caption{The \textbf{Prior Guided Depth Calibration (PGDC)} module leverages 2D detection boxes as priors to precisely target and correct the most severe feature misalignments, which are caused by calibration errors and motion distortion. By applying localized smoothing to the point cloud within these detected regions, the module corrects the erroneous depth information. Simultaneously, it enhances the features of these critical image areas.}
\label{fig3}
\end{figure*}
\section{Related Work}
\textbf{LiDAR-Only 3D Object Detection.}
LiDAR-based 3D detection methods are categorized by their data representation. Point-based methods \cite{qi2017pointnet++,qi2018frustum,li2021lidar} process raw point clouds with MLPs. Voxel-based methods \cite{liu2024sparsedet,chen2023largekernel3d,zhou2018voxelnet} use sparse 3D convolutions on discretized grids, with efficient pillar-based variants \cite{lang2019pointpillars} using 2D backbones. Point-voxel hybrid methods \cite{song2023vp,miao2021pvgnet} combine both for higher accuracy at the cost of greater computational overhead.

\noindent\textbf{Camera-Only 3D Object Detection.}
Camera-based 3D detection has shifted toward multi-view systems \cite{zhang2025geobev,wang2022detr3d}, which outperform monocular methods \cite{lu2021geometry,pu2025monodgp} but increase computational complexity. The Lift-Splat-Shoot (LSS) \cite{philion2020lift} paradigm, which projects image features into 3D using depth estimation, is a key development. This has inspired frameworks that use LiDAR for depth supervision \cite{reading2021categorical}, such as BEVDet \cite{huang2022bevdet4d}, and other techniques that distill LiDAR information \cite{guo2025promptdet}.

\noindent\textbf{LiDAR-Camera Fusion 3D Object Detection.}
Multi-modal fusion is now the standard for 3D object detection. Fusion strategies have evolved from early point-level methods that augment raw LiDAR points with image features \cite{yin2021multimodal,liu2022epnet++,wang2021pointaugmenting}, to more advanced feature-level approaches that use attention mechanisms to integrate 3D proposals with image features \cite{song2024robofusion,song2023graphalign,chen2022deformable}. The current state-of-the-art primarily uses BEV-based fusion \cite{cai2023objectfusion,ge2023metabev,liang2022bevfusion,song2024graphbev}, which unifies both modalities in a shared bird's-eye-view space for more efficient and robust cross-modal interaction.

\section{Method}
\subsection{Framework Overview}
As shown in \cref{fig2}, our proposed framework is a multi-sensor fusion pipeline built upon the strong BEVFusion \cite{liang2022bevfusion} baseline, designed to explicitly mitigate feature misalignment in Bird's Eye View (BEV) perception. The framework processes inputs from a set of $N$ surround-view cameras and a 360-degree LiDAR sensor.
Our contribution is a combination of three synergistic modules. First, \textbf{Prior Guided Depth Calibration (PGDC)} uses 2D bounding boxes to correct misaligned LiDAR points at object boundaries, outputting a refined sparse depth map and enhanced image features. Second, \textbf{Discontinuity Aware Geometric Fusion (DAGF)} uses the corrected depth to generate a dense representation that captures reliable geometric structure.
Finally, the \textbf{Structural Guidance Depth Modulator (SGDM)} fuses the enhanced image features (from PGDC) and the dense geometric representation (from DAGF) to predict a depth distribution for each camera view. Following the Lift-Splat-Shoot \cite{philion2020lift} paradigm, these are projected into a unified BEV feature map, which is then fused with LiDAR BEV features to produce robust 3D detection results.

\subsection{Prior Guided Depth Calibration (PGDC)}
As shown in \cref{fig3}, the Prior Guided Depth Calibration (PGDC) module operates independently on each of the $N$ camera views to correct the initial sparse depth supervision derived from LiDAR. For each view $i \in \{1,..., N\}$, the inputs are the image features $F_{\text{img}}^{(i)} \in \mathbb{R}^{H \times W \times C}$ and the corresponding sparse depth map $D_{\text{raw}}^{(i)} \in \mathbb{R}^{H \times W}$, which is generated by projecting the LiDAR point cloud into that camera's image plane. To better explain this module, we have divided it into the \textbf{2D Guided Depth Align Module} and the \textbf{2D Camera Features Enhance Module}.

First, a 2D detection head provides a set of bounding boxes $\{B_j^{(i)}\}$ for each image. In \textbf{2D Guided Depth Align Module}, based on our observation that misalignment is concentrated at object boundaries, we use these boxes to isolate critical regions. For each box $B_j^{(i)}$, we filter the LiDAR points whose projections fall within it. To mitigate misalignment, we introduce a novel smoothing operation. Instead of simple averaging, our method captures the local depth structure more effectively. For each LiDAR point projected to a pixel $p$ with depth $d_p$, we first use a KD-Tree to find its 10 nearest neighbors, $\mathcal{N}_p$. We then select the 2 neighbors with the smallest depth and the 2 with the largest depth, forming a set of four critical neighbors, $\mathcal{N}_{\text{critical}} \subset \mathcal{N}_p$. This selection strategy is designed to simultaneously capture the object's own depth consistency (via the nearest points) and the sharp depth discontinuity at the boundary between the object and the background (via the farthest points), thereby preserving critical information while smoothing noise. The depth of the original point is concatenated with the depths of these four selected neighbors, creating a 5-channel feature map, $f_p$, for that pixel:
\begin{equation}
f_p = \text{concat}(d_p, \{d_q\}_{q \in \mathcal{N}_{\text{critical}}})
\end{equation}

This feature map is then processed through a lightweight convolutional block to produce the final single-channel, smoothed depth value:
\begin{equation}
d'_{\text{aligned}}(p) = \text{ReLU}(\text{BN}(\text{Conv}(f_p)))
\end{equation}

This process corrects erroneous depth values, resulting in a refined sparse depth map, $D_{\text{aligned}}^{(i)}$.


Simultaneously, in \textbf{2D Camera Features Enhance Module}, we enhance the image features within these critical regions. For each bounding box $B_j^{(i)}$ with a predicted class label $k$, the corresponding image features are amplified by a class-specific hyperparameter $\alpha_k$. This enhancement is applied to all pixels $p$ within the bounding box $B_j^{(i)}$ and across all $C$ feature channels. The operation is defined as:
\begin{equation}
F_{\text{enhanced}}(p,c) = \alpha_k \cdot F_{\text{img}}(p,c)
\end{equation}
where the value of $\alpha_k$ is a class-specific hyperparameter, set based on the object's typical size. The underlying principle is that smaller objects require a stronger feature boost to ensure their representation is not neglected during fusion. Consequently, small classes like pedestrians and traffic cones are assigned a higher $\alpha_k$, whereas larger classes such as buses and trucks receive a more moderate value. This allows for a more targeted enhancement, tailoring the amplification intensity to the specific characteristics of each object category.

These enhanced features are then processed through a Squeeze-and-Excitation (SE) block to adaptively recalibrate channel-wise feature responses, producing the final enhanced image features $F_{\text{enhanced}}^{(i)}$. This ensures that the network can learn to dynamically emphasize more informative channels for each class-specific enhancement.

\begin{figure*}[t]
\centering
\includegraphics[width=1\textwidth]{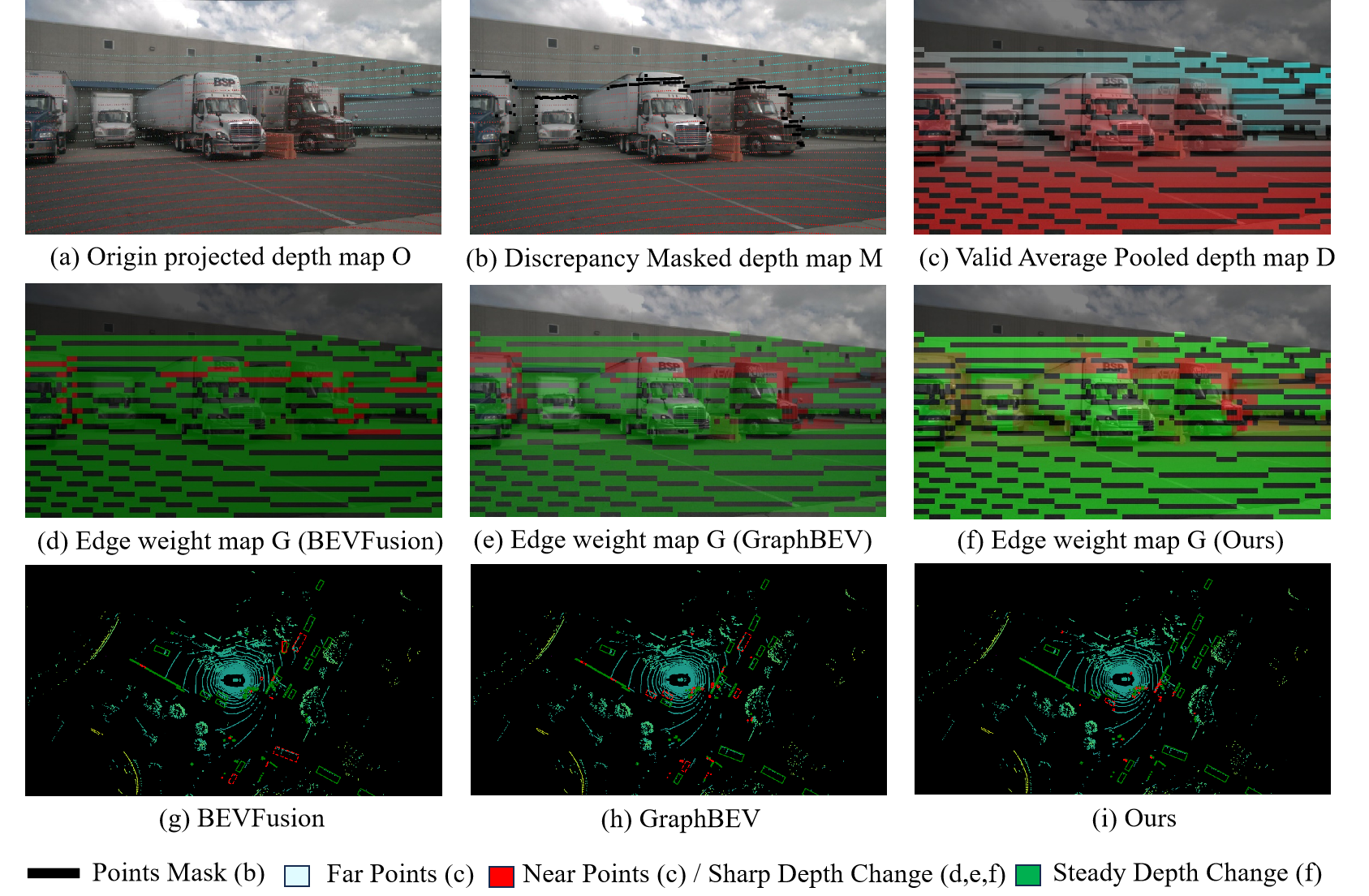} 
\caption{(a) Represents the original projected depth. (b) Represents the projected depth after applying Discrepancy Masking.
(c) Shows the block-wise depth map after Block-based Densification;
(d), (e), and (f) are the final depth change magnitude maps of different methods after Block-based Gradient Extraction. (g), (h) and (i) are visualization of detection results, in which green boxes are True Positives (TP), solid red boxes are False Positives (FP), and dashed red boxes are False Negatives (FN). It is obvious that our method (f) outperforms GraphBEV (e) and BEVFusion (d) because we accurately delineated regions with drastic depth variations and avoided over-smoothing.}
\label{fig:comp}
\end{figure*}

\subsection{Discontinuity Aware Geometric Fusion (DAGF)}
The Discontinuity Aware Geometric Fusion (DAGF) module is engineered to generate a dense and structurally aware depth representation, which serves as a guide for the final depth estimation. For each camera view $i$, it utilizes two inputs: the original sparse depth map, denoted as $D_{\text{raw}}^{(i)}$, and a refined sparse depth map, $D_{\text{aligned}}^{(i)}$, which is produced by the Prior Guided Depth Calibration (PGDC) module. The module's operation is structured as a sequential pipeline.

First, to mitigate noise via \textbf{Discrepancy Masking}, a discrepancy map $\Delta^{(i)}$ is computed. This map is defined as the absolute difference between the raw and aligned depth maps:
\begin{equation}
\Delta^{(i)} = |D_{\text{raw}}^{(i)} - D_{\text{aligned}}^{(i)}|
\end{equation}

Pixels where this discrepancy surpasses a threshold $\tau$, defined as 10\% of the pixel's value in $D_{\text{raw}}^{(i)}$, are considered unreliable and subsequently masked out. This process yields a cleaner, but still sparse, depth map, $M^{(i)}$, where the value of a pixel at coordinates $(u, v)$ is given by:
\begin{equation}
M^{(i)}(u, v) = 
\begin{cases} 
D_{\text{aligned}}^{(i)}(u, v) & \text{if } \Delta^{(i)}(u, v) \leq \tau \\ 
0 & \text{if } \Delta^{(i)}(u, v) > \tau 
\end{cases}
\end{equation}

Next, we perform \textbf{Block-based Densification and Gradient Extraction}. The sparse map $M^{(i)}$ is divided into non-overlapping $20 \times 20$ blocks. For each block, we compute two key statistics from the valid (non-zero) points within it:
\begin{enumerate}
    \item \textbf{Average Depth ($d_{\text{avg}}$):} The mean of all valid depth values in the block. This captures the block's general distance and is used for densification.
    \item \textbf{Maximum Gradient ($g_{\max}$):} The maximum local depth discontinuity within the block. This is found by first calculating an individual gradient for each point in the block (as the max depth difference to its neighbors), and then taking the maximum of these individual gradients.
\end{enumerate}

These two statistics are then broadcast to all pixels within their respective blocks, creating a densified depth map $D_{\text{dense}}^{(i)}$ and a densified gradient map $G_{\text{dense}}^{(i)}$. The visualization of these maps is shown in \cref{fig:comp}, in which we also show the visualization of comparison between our method and others. Our method effectively addresses the misalignment between LiDAR and camera images. Furthermore, in regions with gradual depth changes, it successfully preserves correct depth information while GraphBEV \cite{song2024graphbev} falsely modifies already-correct depth values.

The ultimate output of the \textbf{Discontinuity Aware Geometric Fusion (DAGF)} module for each view is a multi-channel feature map, $F_{\text{FA}}^{(i)} \in \mathbb{R}^{H \times W \times 2}$. This feature map is formed by concatenating the two dense maps:
\begin{equation}
F_{\text{FA}}^{(i)} = [D_{\text{dense}}^{(i)} \oplus G_{\text{dense}}^{(i)}]
\end{equation}
where $\oplus$ denotes concatenation along the channel dimension. 
This combined representation provides both smoothed depth information and explicit boundary-aware structural cues to the subsequent \textbf{Structural Guidance Depth Modulator (SGDM)}.

The features generated by Discontinuity Aware Geometric Fusion (DAGF) guide final depth prediction, $\hat{D}^{(i)}$, from SGDM. The training is supervised by a composite loss function that leverages both the dense map and the gradient map.

\textbf{Focal Loss.} We use the densified map $D_{\text{dense}}^{(i)}$ as direct supervision for the predicted depth $\hat{D}^{(i)}$. The Focal Loss, $\mathcal{L}_{\text{focal}}$, is the average of a per-pixel loss term, $l_{\text{focal}}(u,v)$, over all valid pixels:
\begin{equation}
\mathcal{L}_{\text{focal}} = \frac{1}{|\mathcal{V}|} \sum_{(u,v) \in \mathcal{V}} l_{\text{focal}}(u,v)
\end{equation}

The per-pixel term is defined as $l_{\text{focal}}(u,v) = \text{FL}(\hat{D}^{(i)}(u,v), D_{\text{dense}}^{(i)}(u,v))$. Here, $\mathcal{V}$ is the set of all pixel coordinates with a valid (non-zero) depth, and $\text{FL}(\cdot, \cdot)$ denotes the Focal Loss function. The hyperparameters for the Focal Loss function are set to $\gamma = 2.0$ and $\alpha = 0.25$.

\textbf{Edge-Critical Loss.} To enforce sharp structural boundaries, the \textbf{Edge-Critical Loss}, $\mathcal{L}_{\text{edge}}$, reuses the per-pixel term $l_{\text{focal}}(u,v)$ from the Focal Loss. It introduces a weight from the gradient map, $G^{(i)}(u,v)$, to amplify the penalty at depth discontinuities:
\begin{equation}
\mathcal{L}_{\text{edge}} = \frac{1}{|\mathcal{V}|} \sum_{(u,v) \in \mathcal{V}} G^{(i)}(u,v) \cdot l_{\text{focal}}(u,v)
\end{equation}

This formulation compels the network to prioritize accuracy in regions critical to structural integrity. The final training objective is a composite loss, combining these two depth-specific losses with the standard classification and bounding box regression losses:
\begin{equation}
\mathcal{L}_{\text{total}} = \mathcal{L}_{\text{focal}} + \mathcal{L}_{\text{edge}} + \mathcal{L}_{\text{cls}} + \mathcal{L}_{\text{box}}
\end{equation}

\begin{table*}[htbp]
\centering
\footnotesize
\caption{Performance comparison of 3D object detection methods on the nuScenes validation set across 10 classes: Car (C), Truck (T), Construction Vehicle (CV), Bus (B), Trailer (Tr), Barrier (Ba), Motorcycle (M), Bicycle (Bi), Pedestrian (P), and Traffic Cone (TC).}
\label{tab:3d-detection}
\begin{tabular}{@{} l c cc *{10}{c} @{}}
\toprule
Method & Present at & mAP & NDS & C & T & CV & B & Tr & Ba & M & Bi & P & TC \\
\midrule
TransFusion-L \cite{bai2022transfusion} & CVPR 22 & 65.5 & 70.2 & 86.2 & 54.8 & 26.5 & 70.1 & 42.3 & 72.1 & 69.8 & 53.9 & 86.4 & 70.3 \\
SAFDNet \cite{zhang2024safdnet} & CVPR 24 & 66.3 & 71.0 & 87.6 & 60.8 & 26.6 & 78.0 & 43.5 & 69.7 & 75.5 & 58.0 & 87.8 & 75.0 \\
BEVFusion-PKU \cite{liang2022bevfusion} & NeurIPS 22 & 67.9 & 71.0 & 87.3 & 59.8 & 28.9 & 73.5 & 41.2 & 73.8 & 74.6 & 59.8 & 85.4 & 68.2 \\
LION-Mamba \cite{liu2024lion} & NeurIPS 24 & 68.0 & 72.1 & 87.9 & 64.9 & 28.5 & 77.6 & 44.4 & 71.6 & 75.6 & 59.4 & 89.6 & 80.8 \\
FSHNet \cite{liu2025fshnet} & CVPR 25 & 68.1 & 71.7 & 88.7 & 61.4 & 26.3 & 79.3 & 47.8 & 72.3 & 76.7 & 60.5 & 89.3 & 78.6 \\
BEVFusion-MIT \cite{liu2023bevfusion} & ICRA 23 & 68.5 & 71.4 & 88.2 & 61.7 & 30.2 & 75.1 & 41.5 & 72.5 & 76.3 & 64.2 & 87.5 & 81.0 \\
UniMamba \cite{jin2025unimamba} & CVPR 25 & 68.5 & 72.6 & 88.7 & 64.7 & 28.7 & \textbf{79.7} & \textbf{47.9} & 72.3 & 74.6 & 59.1 & 89.7 & 79.5 \\
M3Net \cite{chen2025m3net} & AAAI 25 & 69.0 & 72.4 & 89.0 & 64.5 & 30.3 & 77.9 & 47.5 & 73.2 & 76.5 & 61.4 & 89.2 & 80.4 \\
BEVDiffuser \cite{ye2025bevdiffuser} & CVPR 25 & 69.2 & 71.9 & 88.5 & 63.5 & 31.0 & 75.3 & 46.2 & 73.2 & 77.5 & 62.8 & 87.9 & 80.5 \\
GraphBEV \cite{song2024graphbev} & ECCV 24 & 70.1 & 72.9 & \textbf{89.8} & 64.2 & 31.2 & 75.8 & 43.5 & 75.6 & 79.3 & 66.3 & 88.6 & 80.9 \\
Ours & - & \textbf{71.5} & \textbf{73.6} & \textbf{89.8} & \textbf{68.5} & \textbf{35.1} & 77.2 & 45.5 & \textbf{78.0} & \textbf{80.5} & \textbf{68.3} & \textbf{90.1} & \textbf{82.0} \\
\bottomrule
\end{tabular}
\end{table*}
\begin{table*}[htbp]
\centering
\caption{Performance comparison of 3D object detection methods on the Argoverse 2 validation set. The best result in each column is highlighted in \textbf{bold}.}
\label{tab:detailed_map_comparison_transposed}
\resizebox{\textwidth}{!}{%
\begin{tabular}{l ccccccccccccccccccccccc}
\toprule
Method & \rotatebox{90}{mAP} & \rotatebox{90}{Vehicle} & \rotatebox{90}{Bus} & \rotatebox{90}{Pedestrian} & \rotatebox{90}{StopSign} & \rotatebox{90}{BoxTruck} & \rotatebox{90}{Bollard} & \rotatebox{90}{C-Barrel} & \rotatebox{90}{Motorcyclist} & \rotatebox{90}{MPC-Sign} & \rotatebox{90}{Motorcycle} & \rotatebox{90}{Bicycle} & \rotatebox{90}{A-Bus} & \rotatebox{90}{SchoolBus} & \rotatebox{90}{TruckCab} & \rotatebox{90}{C-Cone} & \rotatebox{90}{V-Trailer} & \rotatebox{90}{Sign} & \rotatebox{90}{LargeVehicle} & \rotatebox{90}{Stroller} & \rotatebox{90}{Bicyclist} & \rotatebox{90}{Truck} & \rotatebox{90}{Dog} \\
\midrule
VoxelNeXt \cite{chen2023voxelnext} & 30.7 & 65.2 & 40.1 & 42.5 & 64.9 & 46.8 & 40.1 & 52.3 & 22.5 & 38.0 & 24.1 & 21.3 & 56.2 & 56.0 & 39.9 & 38.1 & 54.8 & 20.2 & 38.0 & 18.2 & 27.1 & 36.3 & 11.5 \\
BEVFusion \cite{liang2022bevfusion} & 38.1 & 72.1 & 45.3 & 48.0 & 73.5 & 51.2 & 43.5 & 58.9 & 25.8 & 42.1 & 27.3 & 23.9 & 61.1 & 61.3 & 45.2 & 42.0 & 58.2 & 23.5 & 41.9 & 21.0 & 30.2 & 40.1 & 14.8 \\
SAFDNet \cite{zhang2024safdnet} & 39.7 & 74.3 & 46.8 & 49.5 & 75.1 & 53.0 & 45.1 & 60.3 & 27.8 & 43.5 & 28.5 & 25.0 & 63.2 & 62.9 & 47.8 & 43.1 & 60.1 & 24.9 & 43.2 & 22.5 & 31.8 & 41.9 & 16.2 \\
FSHNet \cite{liu2025fshnet} & 40.2 & 75.1 & 47.1 & 50.3 & 76.2 & 54.9 & 46.0 & 62.1 & 28.5 & 45.8 & 29.1 & 25.4 & 64.8 & 64.1 & 48.9 & 44.5 & 61.3 & 26.0 & 44.1 & 23.6 & 32.5 & 43.0 & 17.1 \\
LION \cite{liu2024lion} & 40.7 & 74.7 & 47.5 & 57.1 & 77.0 & 55.1 & 48.3 & 63.7 & 29.7 & 47.3 & 27.0 & 25.2 & 66.9 & 63.7 & 44.2 & 42.5 & 57.9 & 22.0 & 39.3 & 19.9 & 28.8 & 37.7 & 12.8 \\
M3Net \cite{chen2025m3net} & 40.9 & 74.9 & 47.8 & 57.4 & 77.1 & 55.3 & 48.5 & 63.9 & 29.9 & 47.5 & 28.2 & 25.4 & 67.0 & 64.2 & 46.8 & 43.8 & 59.5 & 24.0 & 41.5 & 21.8 & 30.5 & 40.2 & 15.2 \\
GraphBEV \cite{song2024graphbev} & 41.1 & 75.2 & 48.0 & 57.6 & 77.3 & 55.7 & 48.8 & 64.2 & 30.1 & 47.7 & 29.5 & 25.8 & \textbf{68.5} & 64.7 & 48.2 & 44.5 & \textbf{62.7} & 26.2 & 44.2 & 23.8 & 32.6 & \textbf{44.2} & 17.2 \\
\textbf{Ours} & \textbf{41.7} & \textbf{76.2} & \textbf{49.0} & \textbf{58.5} & \textbf{78.1} & \textbf{56.8} & \textbf{49.8} & \textbf{65.1} & \textbf{31.2} & \textbf{48.5} & \textbf{31.0} & \textbf{26.9} & 68.1 & \textbf{65.8} & \textbf{50.3} & \textbf{46.0} & 62.5 & \textbf{27.5} & \textbf{45.8} & \textbf{25.0} & \textbf{33.8} & 43.7 & \textbf{18.8} \\
\bottomrule
\end{tabular}
}
\end{table*}
\subsection{Structural Guidance Depth Modulator(SGDM)}

As shown in \cref{fig2}(c), our Structural Guidance Depth Modulator (SGDM) is designed for multi-modal depth estimation, intelligently integrating visual features from the camera with geometric data from our processed depth representation. The architecture first processes camera and depth features through parallel convolutional layers in order to extract and normalize modality-specific features. These encoded features are then concatenated and fed into a processing block where a gated attention mechanism generates a spatial attention map. This map modulates the initial depth prediction, effectively learning a confidence score for each pixel's placement in 3D space.

Recognizing that the fusion process can dilute the rich semantic information inherent in the camera features, we introduce a crucial residual connection to preserve the original camera feature stream. This connection acts as a direct information pathway, carrying the pristine visual features forward and bypassing the fusion block.
This architecture thus creates a powerful synergy: the residual path guarantees the preservation of critical visual context, while the attention gate intelligently modulates the feature map, allowing the network to selectively emphasize reliable information. The final output of this module is a discrete probability distribution over a set of predefined depth bins for each pixel, framing depth estimation as a more stable, per-pixel classification task.

\section{Experiments}
\subsection{Dataset and Metrics}
Our experiments utilize two large-scale, multimodal autonomous driving datasets. The first is nuScenes \cite{caesar2020nuscenes}, which contains 1,000 diverse urban scenes (~20 seconds each) from Boston and Singapore. It includes 1.4M camera images, 390K LiDAR sweeps, 1.4M RADAR sweeps, and 1.4M annotated 3D bounding boxes spanning 23 object classes, along with rich attributes and HD maps. For this benchmark, we evaluate using its main metrics: mean Average Precision (mAP) and nuScenes Detection Score (NDS). We also conduct experiments on Argoverse 2 \cite{wilson2023argoverse}, a perception-focused dataset featuring 1,000 scenarios from six distinct U.S. cities. It provides data from seven cameras and two LiDAR sensors, totaling 700K front-camera images and 20M annotated 3D bounding boxes across 30 object classes. For evaluation on this dataset, we follow its official protocol, primarily using the mAP metric.

\subsection{Implementation Details}
The LiDAR branch utilizes TransFusion-L \cite{bai2022transfusion} for feature encoding to generate Bird's Eye View (BEV) features. Simultaneously, the camera branch processes input images resized and cropped to 448×800 resolution through a Swin Transformer backbone \cite{liu_lin_cao_hu_wei_zhang_lin_guo_2021} with head counts of 3, 6, 12, and 24, followed by multi-scale feature fusion using FPN. For 2D object detection, a YOLOv9 \cite{wang2024yolov9} head is implemented. This combination provides a highly efficient and accurate detector, making it suitable for generating robust 2D priors. The LSS \cite{philion2020lift} configuration defines frustum ranges with X: [-54m, 54m, 0.3m], Y: [-54m, 54m, 0.3m], Z: [-10m, 10m, 20m], and depth: [1m, 60m, 0.5m]. We implement our network in PyTorch, training it on 8 RTX 4090 GPUs. Latency is measured on RTX 4090 GPU. Detailed hyperparameter configurations can be found in Appendix B.

\begin{table}[htbp]
\centering
\caption{Ablation study of our three proposed modules on nuscenes dataset. LT is an abbreviation for latency.}
\label{tab:ablation_final}
\setlength{\tabcolsep}{3.5pt}
\begin{tabular}{ccc|ccc}
\toprule
\textbf{PGDC} & \textbf{DAGF} & \textbf{SGDM} & \textbf{mAP (\%)} & \textbf{NDS (\%)} & \textbf{LT (ms)} \\
\midrule
$\times$ & $\times$ & $\times$ & 67.9 & 71.0 & +0.0 \\
$\checkmark$ & $\times$ & $\checkmark$ & 69.8 & 72.5 & +13.0 \\
$\times$ & $\checkmark$ & $\checkmark$ & 69.0 & 71.6 & +7.0 \\
$\checkmark$ & $\checkmark$ & $\checkmark$ & \textbf{71.5} & \textbf{73.6} & +15.0 \\
\bottomrule
\end{tabular}
\end{table}
\begin{table}[htbp]
\centering
\caption{Granular ablation study. We separate PGDC into DAM: Depth Align Module and FEM: Feature Enhance Module.}
\label{tab:granular_ablation}
\setlength{\tabcolsep}{5.5pt}
\begin{tabular}{cccc|cc}
\toprule
\textbf{DAM} & \textbf{FEM} & \textbf{$D_{\text{dense}}$} & \textbf{$G_{\text{dense}}$} & \textbf{mAP (\%)} & \textbf{NDS (\%)} \\
\midrule
$\times$ & $\times$ & $\times$ & $\times$ & 67.9 & 71.0 \\
\checkmark & $\times$ & $\times$ & $\times$ & 69.4 & 72.1 \\
\checkmark & \checkmark & $\times$ & $\times$ & 69.8 & 72.5 \\
\checkmark & \checkmark & \checkmark & $\times$ & 70.8 & 73.1 \\
\checkmark & \checkmark & \checkmark & \checkmark & \textbf{71.5} & \textbf{73.6} \\
\bottomrule
\end{tabular}
\end{table}
\begin{table}[htbp]
\centering
\caption{Ablation study on the impact of 2D detectors.}
\label{tab:2d_quality_ablation}
\setlength{\tabcolsep}{5.5pt}
\begin{tabular}{lcc}
\toprule
\textbf{2D Prior Source} & \textbf{mAP (\%)} & \textbf{NDS (\%)} \\
\midrule
Random Priors & 68.5 & 71.2 \\
No 2D Priors & 69.0 & 71.6 \\
Full-Image Prior & 69.4 & 71.8 \\
YOLO-X Priors & 70.3 & 72.5 \\
\textbf{Realistic 2D Priors (YOLOv9)} & 71.5 & 73.6 \\
Ground Truth 2D Priors & 73.5 & 74.2 \\
\bottomrule
\end{tabular}
\end{table}

\begin{table}[htbp]
\centering
\caption{Impact of different 2D prior qualities on PGDC module configurations.}
\label{tab:worst_case_ablation}
\begin{tabular}{llcc}
\toprule
\textbf{Module } & \textbf{2D Prior Source} & \textbf{mAP (\%)} & \textbf{NDS (\%)} \\
\midrule
\multirow{3}{*}{DAM } & No Priors & 69.0 & 71.6 \\
& Random Priors & 69.1 & 71.8 \\
& Full-Image Prior & 69.9 & 72.0 \\
\midrule
\multirow{3}{*}{Full Model} & No Priors & 69.0 & 71.6 \\
& Random Priors & 68.5 & 71.2 \\
& Full-Image Prior & 69.4 & 71.8 \\
\midrule
\multirow{3}{*}{FEM } & No Priors & 69.0 & 71.6 \\
& Random Priors & 67.4 & 70.1 \\
& Full-Image Prior & 68.6 & 71.3 \\
\bottomrule
\end{tabular}
\end{table}
\subsection{Comparison Results}
We evaluate our proposed framework against a comprehensive set of recent state-of-the-art (SOTA) 3D object detection methods on two challenging, large-scale benchmarks: nuScenes and argoverse2, as detailed in \cref{tab:3d-detection} and \cref{tab:detailed_map_comparison_transposed}. Our method achieves a new state-of-the-art performance on the nuScenes validation set, reaching \textbf{71.5\% mAP} and \textbf{73.6\% NDS} with negligible increment of inference time. This result surpasses previous leading methods, including GraphBEV \cite{song2024graphbev} (70.1\% mAP, 72.9\% NDS), BEVDiffuser \cite{ye2025bevdiffuser} (69.2\% mAP, 71.9\% NDS), and the strong BEVFusion-PKU \cite{liang2022bevfusion} baseline (67.9\% mAP, 71.0\% NDS). 

\subsection{Ablation Study}
We adopt BEVFusion-PKU \cite{liang2022bevfusion} as baseline to evaluate our proposed modules on nuscenes dataset. As shown in \cref{tab:ablation_final}, each module individually improves performance. Our DAGF module relies on PGDC; for the DAGF-only ablation, we modified it to use the original depth map directly. The results reveal a strong synergistic effect between PGDC and DAGF, where their combined gain exceeds the sum of their individual contributions.
For more specific lantency analysis, please refer to Appendix C.

To further analyze component contributions, \cref{tab:granular_ablation} presents a granular ablation. We incrementally test PGDC's functions, the Depth Align Module (DAM) and the Feature Enhance Module (FEM), and DAGF's representations ($D_{\text{dense}}$ and $G_{\text{dense}}$). The results show DAM provides the most significant initial boost. FEM offers a further gain, followed by the dense depth representation. Finally, the gradient representation pushes the model to its peak performance.

Finally, to isolate the impact of 2D detector accuracy, \cref{tab:2d_quality_ablation} shows an ablation study on 2D prior quality and \cref{tab:worst_case_ablation} gives a deeper insight on how 2D priors affect specific PGDC modules. The results demonstrate that while performance generally scales with the quality of the 2D priors, our method is notably robust. Even with completely random priors, the model's performance is not significantly harmed. Furthermore, using a coarse prior that covers the full image still brings an improvement over using no prior at all. 
The system maintains structural accuracy even when faulty 2D priors cause the PGDC to falsely over-smooth the correct depth. This is because after the DAGF module removes the incorrectly smoothed points, refilling the boundary regions accurately preserves the point cloud's structure.

\section{Conclusion}
In this work, we address the critical issue of LiDAR-camera feature misalignment, most severe at object-background boundaries due to depth-dependent projection errors. To proactively correct these errors stemming from calibration and motion, our framework introduces three novel components. PGDC leverages 2D detection priors to correct depth information in critical regions. DAGF then creates a dense, structurally aware representation. Finally, the SGDM intelligently fuses visual and geometric cues to predict a highly accurate depth distribution for view transformation, effectively mitigating artifacts in the final BEV space.
\clearpage

\section*{Acknowledgements}
This work was supported in part by the Anhui Engineering Research Center for Intelligent Driving Technology and Application, the Jianghuai Advance Technology Center, and the Key Science \& Technology Project of Anhui Province under 
Grant Nos. 202523f12050009 and 202523j08050018.
{
    \small
    \bibliographystyle{ieeenat_fullname}
    \bibliography{AnonymousSubmission/LaTeX/aaai2026}
}


\end{document}


\maketitle

\section*{Supplementary}
\subsection*{A Details about misalignment}

\begin{figure*}[t]
\centering
\includegraphics[width=1\textwidth]{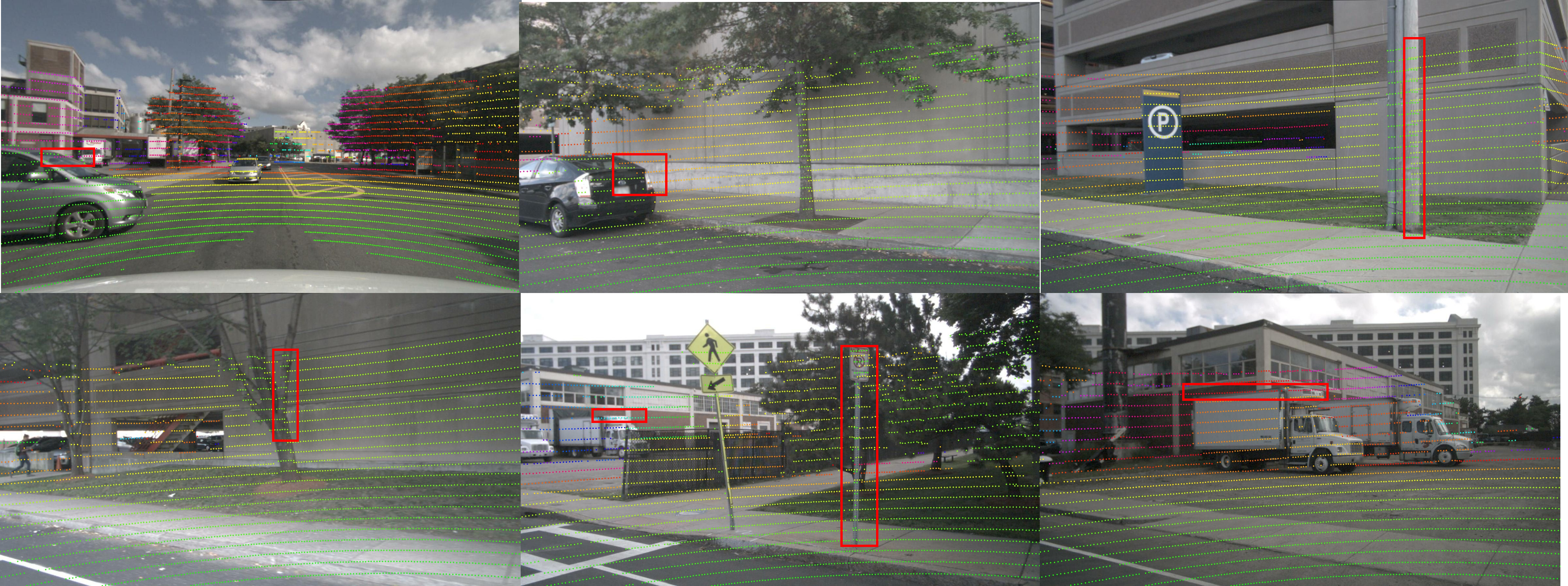} 
\caption{Spatial Misalignment (Extrinsic Calibration Error) and Temporal Misalignment (Synchronization Error) cases in nuscenes dataset}
\label{fig5}
\end{figure*}

The fusion of data from multiple sensor modalities, particularly LiDAR and camera, is a cornerstone of
modern 3D perception systems for autonomous vehicles. Datasets such as nuScenes provide a rich source of multi-modal data to train and validate these systems. However, a critical challenge that persists is the inherent spatial misalignment between the sensor streams. When projecting LiDAR point clouds onto corresponding camera images, a noticeable discrepancy often appears. This artifact is not uniform across the image; instead, it is most pronounced at the boundaries between foreground objects and the distant background. A canonical example is the projection of LiDAR points onto a vehicle, where points corresponding to a far-away background wall erroneously ``bleed" onto the pixels of the much closer vehicle, creating a fringe of incorrectly projected points along the object's silhouette. This document provides a detailed analysis of the primary causes of this phenomenon within the context of the nuScenes dataset.

The core of multi-modal fusion lies in the geometric projection of a 3D point from the LiDAR's coordinate frame ($L$) into the 2D pixel coordinate frame of the camera ($C$). In an ideal system, this is a deterministic transformation. For the nuScenes dataset, this involves transforming a point $P_L \in \mathbb{R}^3$ from the LiDAR frame to the ego-vehicle frame ($E$), and then from the ego-vehicle frame to the camera frame. The extrinsic transformation from the LiDAR to the ego-vehicle is represented by a rigid body transformation matrix $T_{E \leftarrow L} \in SE(3)$, and similarly from the camera to the ego-vehicle, $T_{E \leftarrow C}$. Therefore, the transformation from LiDAR to camera is given by $T_{C \leftarrow L} = (T_{E \leftarrow C})^{-1} \cdot T_{E \leftarrow L}$. This matrix can be decomposed into a rotation matrix $R_{C \leftarrow L}$ and a translation vector $t_{C \leftarrow L}$. A point $P_L$ is projected into the camera's image plane to pixel coordinates $p \in \mathbb{R}^2$ by first converting to homogeneous coordinates $\tilde{P}_L = [P_L; 1]$ and applying the full projection model:
\begin{equation}
z \begin{bmatrix} p \\ 1 \end{bmatrix} = K \begin{bmatrix} R_{C \leftarrow L} | t_{C \leftarrow L} \end{bmatrix} \tilde{P}_L
\end{equation}

where $K \in \mathbb{R}^{3 \times 3}$ is the camera's intrinsic calibration matrix and $z$ is the point's depth in the camera frame. This equation represents the ideal geometric relationship, but in practice, several sources of error corrupt this process.

The first major source of error is static and stems from imperfect extrinsic calibration. The matrix $T_{C \leftarrow L}$ is estimated through a calibration procedure that, while precise, inevitably contains small, residual errors. Let the estimated, erroneous rotation matrix be $\hat{R} = R_{C \leftarrow L} \cdot \Delta R$, where $\Delta R$ represents a small angular error. The effect of this rotational error on the final projection is amplified by the depth of the 3D point. For a small angular error $\Delta\theta$ about an axis, the resulting displacement $\Delta x$ in the projection plane at a depth $d$ can be approximated by:
\begin{equation}
\Delta x \approx d \cdot \tan(\Delta\theta)
\end{equation}
This relationship is fundamental to understanding the boundary-specific nature of the misalignment. A point on a background wall at a depth $d_{\text{bg}} = 60\text{m}$ will experience a projection shift four times larger than a point on a foreground vehicle at $d_{\text{fg}} = 15\text{m}$ for the exact same calibration error $\Delta\theta$. This creates a significant differential shift between the projected locations of foreground and background points, which becomes visually apparent precisely at the depth discontinuities that define object boundaries.

The second category of errors is dynamic, arising from motion during the data acquisition process. The Velodyne HDL-32E LiDAR used in nuScenes is a mechanical spinning sensor operating at 20 Hz, meaning a full 360-degree scan requires 50 ms to complete. A point cloud is therefore not an instantaneous snapshot of the world; each point is captured at a slightly different time and from a slightly different ego-vehicle pose. This is often referred to as a ``rolling shutter" effect. Although nuScenes provides motion compensation to correct for this ego-motion distortion by transforming all points into the coordinate system of the final timestamp, this compensation often assumes linear motion of the ego-vehicle. During complex maneuvers involving non-linear acceleration or yaw rates, residual distortions remain in the point cloud, causing static objects like straight walls to appear warped.

Furthermore, a more subtle but systematic dynamic error in nuScenes arises from the sensor synchronization and timestamping policy. The camera exposure is triggered when the LiDAR's top laser sweeps across the center of the camera's field of view. This moment defines the image timestamp, $t_{\text{cam}}$. However, the corresponding LiDAR scan is assigned a single timestamp, $t_{\text{lidar}}$, which corresponds to the time when the full 360-degree rotation is completed. This creates a systematic temporal offset $\Delta t = t_{\text{lidar}} - t_{\text{cam}}$, where the LiDAR data is, on average, approximately 25 ms older than the camera image it is paired with. For static scenes, this offset is negligible. For any object in motion with velocity $v_{\text{obj}}$, however, this leads to a physical displacement error in its perceived position:
\begin{equation}
\Delta s = v_{\text{obj}} \cdot \Delta t
\end{equation}
For a vehicle moving at 50 km/h (13.9 m/s), a 25 ms offset results in a spatial error of approximately 35 cm, a significant discrepancy that further contributes to the misalignment between the LiDAR projection and the camera image.

In conclusion, the visually striking misalignment at object-background boundaries in datasets like nuScenes is not a result of a single flaw, but rather the confluence of static, dynamic, and temporal errors. The depth-dependent amplification of minute calibration errors creates a foundational differential shift between near and far objects. This is compounded by residual distortions from imperfect ego-motion compensation and systematic spatial offsets for moving objects due to timestamp asynchrony. The reason these artifacts become so prominent at object boundaries is because these boundaries represent a sharp discontinuity in depth and motion state. At these locations, the a small error vector associated with the foreground object's projection meets the large, dissimilar error vector of the background, making the misalignment manifest as a visible ``bleeding" of points. These effects are visually summarized in \cref{fig5}. Recognizing that these misalignments are an inherent and predictable artifact of the data collection process is crucial for developing robust perception algorithms that can account for, and be resilient to, such real-world data imperfections.

\subsection*{B Hyperparameter Ablation Study}
All performance results, including the primary results in the main text, are averaged over 3 separate runs. To validate our design choices, we conducted additional ablation studies on key hyperparameters within our proposed modules. The results, shown in \cref{tab:hyper_ablation}, demonstrate the robustness of our framework and justify the parameter values used in our main experiments.
\begin{table*}[htbp]
\centering
\caption{Efficiency and performance comparison. Our primary results (71.5 mAP) are based on the bevfusion-pku backbone. When compared to methods on the same \textbf{bevfusion-mit} baseline, our approach is notably faster than GraphBEV with higher mAP and same NDS.}
\label{tab:efficiency_comparison}
\begin{tabular}{lccc}
\toprule
\textbf{Method} & \textbf{mAP (\%)} & \textbf{NDS (\%)} & \textbf{Latency (ms)} \\
\midrule
bevfusion-pku (baseline) & 67.9 & 71.0 & 270 \\
\textbf{Ours (bevfusion-pku based)} & \textbf{71.5} & \textbf{73.6} & \textbf{285} \\
\midrule
bevfusion-mit (baseline) & 68.5 & 71.4 & 120 \\
GraphBEV (bevfusion-mit based) & 70.1 & 72.9 & 142 \\
\textbf{Ours (bevfusion-mit based)} & 71.1 & 72.9 & \textbf{137} \\
\bottomrule
\end{tabular}
\end{table*}
\begin{table}[htbp]
\centering
\caption{Ablation studies for key hyperparameters. The default configuration used in our main experiments is highlighted in bold.}
\label{tab:hyper_ablation}
\begin{tabular}{llcc}
\toprule
\textbf{Module} & \textbf{Hyperparameter} & \textbf{mAP (\%)} & \textbf{NDS (\%)} \\
\midrule
\multirow{3}{*}{DAGF} & Block Size: $10 \times 10$ & 71.2 & 73.3 \\
& Block Size: $\mathbf{20 \times 20}$ & \textbf{71.5} & \textbf{73.6} \\
& Block Size: $40 \times 40$ & 71.0 & 73.2 \\
\midrule
\multirow{3}{*}{DAM} & Neighbors ($K_s$): 5 & 71.1 & 73.4 \\
& Neighbors ($K_s$): \textbf{10} & \textbf{71.5} & \textbf{73.6} \\
& Neighbors ($K_s$): 20 & 71.4 & 73.5 \\
\midrule
\multirow{3}{*}{FEM} & $\alpha_k$: None (disabled) & 71.0 & 72.9 \\
& $\alpha_k$: Shared ($\alpha=1.1$) & 71.2 & 73.2 \\
& $\alpha_k$: \textbf{Class-Specific} & \textbf{71.5} & \textbf{73.6} \\
\bottomrule
\end{tabular}
\end{table}
\begin{table}[htbp]
\centering
\caption{Impact of different 2D prior qualities on module configurations. We test ``No Priors", misleading ``Random Priors", and a ``Full-Image Prior" (a box covering the entire image). Note that the Depth Align Module shows unique resilience to both random and full-image priors, while FEM is sensitive to any incorrect prior.}
\label{tab:worst_case_ablation}
\begin{tabular}{llcc}
\toprule
\textbf{Model } & \textbf{2D Prior Source} & \textbf{mAP (\%)} & \textbf{NDS (\%)} \\
\midrule
\multirow{3}{*}{DAM } & No Priors & 69.0 & 71.6 \\
& Random Priors & 69.1 & 71.8 \\
& Full-Image Prior & 69.9 & 72.0 \\
\midrule
\multirow{3}{*}{Full Model} & No Priors & 69.0 & 71.6 \\
& Random Priors & 68.5 & 71.2 \\
& Full-Image Prior & 69.4 & 71.8 \\
\midrule
\multirow{3}{*}{FEM Only} & No Priors & 69.0 & 71.6 \\
& Random Priors & 67.4 & 70.1 \\
& Full-Image Prior & 68.6 & 71.3 \\
\bottomrule
\end{tabular}
\end{table}

\textbf{DAGF Block Size.} We evaluated the block size used for depth and gradient densification. A smaller block size of $10 \times 10$ is more sensitive to local noise, slightly degrading performance. A larger block size of $40 \times 40$ over-smooths the geometric details, also leading to a performance drop. The chosen $20 \times 20$ block size provides the best balance between capturing geometric structure and robustness to noise.

\textbf{Depth Align Module Neighbors.} We tested the number of neighbors ($K_s$) used in the Structural Depth Smoothing algorithm. While using 5 neighbors already yields good results, increasing the count to 10 further improves performance by capturing a more stable local depth distribution. Increasing to 20 neighbors did not provide significant additional gains and slightly increased computational cost, justifying our choice of $K_s=10$.

\textbf{FEM Feature Enhancement.} We analyzed the impact of the feature enhancement factor $\alpha_k$. Disabling enhancement ($\alpha_k$=None) leads to a significant performance drop, showing the value of boosting features in critical areas. Using a single, shared factor for all classes improves results, but employing class-specific factors (e.g., $\alpha_k=1.4$ for large objects and $\alpha_k=1.8$ for small objects) allows the model to adapt more effectively to the unique characteristics of different object categories, yielding the best performance.

\textbf{Depth Align Module Smoothing Method.} To validate the effectiveness of our proposed structural smoothing method over simpler alternatives, we conducted an ablation study within our full model configuration. We replaced our method with a baseline that performs simple averaging of the depth values of the $K_s$ nearest neighbors. As shown in \cref{tab:smoothing_ablation_supp}, our structural smoothing, which intelligently selects and weights neighbors to preserve depth discontinuities, yields a clear improvement in both mAP and NDS over the simpler method. This confirms that the nuanced approach is critical for achieving the best performance by correcting misalignment without erasing important geometric features at object boundaries.

\begin{table}[h]
\centering
\caption{Ablation on the depth smoothing method. The comparison is performed by swapping the smoothing component within the full model configuration. Our proposed method shows a clear advantage over a simple averaging baseline.}
\label{tab:smoothing_ablation_supp}
\begin{tabular}{lcc}
\toprule
\textbf{Smoothing Method} & \textbf{mAP (\%)} & \textbf{NDS (\%)} \\
\midrule
Simple Averaging & 70.8 & 73.0 \\
\textbf{Structural Smoothing (Ours)} & \textbf{71.5} & \textbf{73.6} \\
\bottomrule
\end{tabular}
\end{table}

\subsection*{C Efficiency Analysis}
We analyzed the computational cost and performance of our method on different backbones. \cref{tab:efficiency_comparison} summarizes the mAP, NDS, and inference latency, comparing our approach with the corresponding baselines and other methods like GraphBEV.

The results show our method's effectiveness. On the \textbf{bevfusion-pku} backbone, we significantly improve mAP (from 67.9 to 71.5) and NDS (from 71.0 to 73.6) with a modest 15 ms overhead. On the \textbf{bevfusion-mit} backbone, our method is 5 ms (142 ms vs 137 ms) faster than GraphBEV while operating on the same baseline.

\subsection*{D Detailed Analysis of 2D Detector Impact}
Our investigation into the influence of 2D detector quality revealed distinct effects on our proposed modules, particularly the Depth Align Module (DAM) and the Feature Enhancement Module (FEM), especially under conditions of extremely poor 2D priors.

For the \textbf{Depth Align Module}, we observed a surprising degree of resilience to poor-quality priors. As detailed in \cref{tab:worst_case_ablation}, the module's performance improves over the baseline even when guided by flawed information. Counter-intuitively, providing completely \textbf{Random Priors} still results in a performance gain (+0.1\% mAP), and using a \textbf{Full-Image Prior} yields an even more significant boost (+0.9\% mAP). This suggests that the Depth Align Module can extract a beneficial signal from almost any spatial guidance, leveraging the general depth distribution within the indicated region—however coarse or inaccurate—to aid the 3D perception task.

In contrast, the \textbf{Feature Enhancement Module (FEM)} exhibited greater sensitivity to the quality of 2D detections. When supplied with incorrect bounding boxes, the FEM's performance could degrade below the baseline. This occurs because the FEM is designed to intensify features within specific, object-centric regions. An incorrect prior causes it to amplify irrelevant or misleading background features while neglecting the actual object, thereby introducing noise that can confuse the final detection head. The quantitative results of this analysis are presented in \cref{tab:worst_case_ablation}.
\subsection*{E Implementation}
The code for our proposed method can be fully reproduced following the descriptions in the main paper's Method section. Our implementation is built upon the official \textbf{bevfusion-pku} baseline. It is important to note that the specific code handling the projection is provided in our appendix file.
And the code of \textbf{Structural Guidance Depth Modulator (SGDM)} is provided in appendix file, while the complete code of this paper will go public upon acceptance.
